\documentclass[10pt]{article}
\usepackage[preprint]{tmlr}
\usepackage{amsmath,amssymb,amsthm}
\usepackage{booktabs}
\usepackage{microtype}
\usepackage{graphicx}
\usepackage[hidelinks]{hyperref}

\newtheorem{theorem}{Theorem}
\newtheorem{lemma}{Lemma}[section]
\newtheorem{corollary}[theorem]{Corollary}
\DeclareMathOperator{\Var}{Var}

\DeclareMathOperator{\Beta}{Beta}
\DeclareMathOperator{\GammaLaw}{Gamma}
\newcommand{\E}{\mathbb{E}}
\newcommand{\authororcid}[1]{\,\href{https://orcid.org/#1}{%
  \raisebox{-.15ex}{\includegraphics[height=1.55ex]{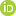}}}}

\title{Corrected mean depths and pairwise score fluctuations in isolation trees}
\author{Sehej Jain\textsuperscript{*}\authororcid{0000-0002-7265-7648},
  Agastya Seth\authororcid{0009-0008-9139-4207}, and Yash Tomar\authororcid{0009-0000-6521-215X}\\
  {\normalfont Independent Researchers\quad\textsuperscript{*}Corresponding author}}

\begin{document}
\maketitle

\begin{abstract}
We study complete one-dimensional isolation trees trained on independent observations from a positive Lipschitz density on a compact interval. The mean held-out depth has additive constant \(2\gamma-\tfrac12\) in the interior and \(\gamma\) at or beyond an endpoint, correcting the asymptotes of Zheng et al. Inserting an interior query raises its mean by half a cut asymptotically. On two separated intervals the gap between components also contributes a finite depth term. For a reference pool of size \(N\), averaging all size-\(m\) subsets and all cuts leaves variance of order \(m/N\) for a fixed query and for the depth difference between two fixed distinct interior queries. When \(m/N\to0\) and \(m\to\infty\), we obtain the leading variance constants from the effect of adding one point to a local Poisson process. Normalizing the depths suppresses absolute fluctuations, while the monotone isolation score retains the same pairwise ordering.
\end{abstract}


\section{Introduction}

Isolation forests score observations by how soon random cuts separate them from a training sample \citep{liu2008,liu2012}. \citet{zheng2026} derive exact finite-sample formulas for complete one-dimensional trees and identify separate density and position terms in the mean depth. We retain the nearest spacings when computing the additive constants.\footnote{Generative AI tools assisted with literature review, mathematical derivations and checks, implementation, numerical analysis, and manuscript drafting. The authors assume responsibility for all content.} These spacings stay random on the scale of their denominators. In particular the two training depths bracketing an interior query can have a nonvanishing difference. Devroye's limit laws for insertion depths in split trees \citep{devroye1998} give a broader context for logarithmic growth. Mass estimation \citep{ting2013} uses random partitions to estimate probability content and provides a related approach to anomaly scoring. Our mean results concern a fixed query in a compact support model. Unbounded laws such as Gaussian or exponential distributions fall outside these assumptions.

Resampling without replacement is a standard form of aggregation \citep{buhlmann2002,buja2006}, and subsampled ensembles admit a \(U\)-statistic representation \citep{hoeffding1948,mentch2016,wager2018,peng2022}. We study how a single observation near a query affects its score across training pools. This links the fluctuations of a pairwise difference to the order-one density and centrality terms that set its population scale.

\begin{itemize}
\item Theorem~\ref{thm:mean} corrects the additive constants for held-out mean depths and gives the asymptotic half-cut effect of inserting an interior query.
\item Corollary~\ref{cor:two} extends the mean calculation to two separated support intervals and identifies the contribution from the gap between them.
\item Theorem~\ref{thm:variance} proves a shared-pool variance rate of \(m/N\) for fixed queries and their pairwise contrasts. It gives explicit Poisson expressions for the leading constants when \(m\to\infty\) and \(m/N\to0\).
\end{itemize}

\section{Depth conventions and finite-sample identities}
\label{sec:model}

For a finite set \(S\) of distinct real numbers, cut each nonsingleton node at a point uniform between its sample minimum and maximum, independently at successive nodes. Grow to singleton leaves and count cuts, with root depth zero. Let \(D(x;S)\) denote the depth of a query \(x\) averaged over the cuts. These conventions agree with Algorithms 1 and 2 of \citet{zheng2026} in one dimension.

For an anchor \(z\in S\), order the positive distances from \(z\) to the points of \(S\) on each side. The depth of \(z\) is the sum of the two side contributions. For a distance list \(d=(d_1,\ldots,d_k)\), the contribution is
\begin{equation}
 \mathcal L(d)=\sum_{j=1}^k\frac{d_j-d_{j-1}}{d_j},\qquad d_0=0.
 \label{eq:leaf}
\end{equation}
An empty list contributes zero. If \(x\) is between consecutive observations \(z_-<z_+\), its depth is
\begin{equation}
 D(x;S)=\lambda D(z_-;S)+(1-\lambda)D(z_+;S),\qquad
 \lambda=\frac{z_+-x}{z_+-z_-}.
 \label{eq:interpolate}
\end{equation}
Queries beyond the sample range inherit the nearest extreme point's depth. These identities are Theorems 3.5 and A.2 in \citet{zheng2026}.

One derivation assigns each gap an independent exponential clock of rate equal to its length, along with an independent uniform cut location in the gap. This is the exponential-priority form of the interval-ancestor calculation for monotone random trees in \citet{morales2023}. At a node the first clock selects a gap with probability proportional to its length. Memorylessness gives the same cut law recursively. A gap to one side of a training point is cut on its path precisely when its clock precedes all clocks between that point and the gap's far end. The probability is the gap length divided by that total distance, giving \eqref{eq:leaf}. The cut within the gap containing a held-out query determines the bracketing leaf reached, independently of the priority tree. This gives \eqref{eq:interpolate}.

Let \(X_1,X_2,\ldots\) be iid with a density \(f\) on \([a,b]\) which is Lipschitz there, with \(0<f_{\min}\le f\le f_{\max}<\infty\). Write \(F\) for its cumulative distribution function and \(S_m=\{X_1,\ldots,X_m\}\). For deterministic \(x\), write \(D_m(x)=D(x;S_m)\), \(\mu_m(x)=\E D_m(x)\), and \(D_m^{\mathrm{ins}}(x)=D(x;S_m\cup\{x\})\). The inserted tree has \(m\) other observations. Counterfactual insertion is also used in robust random cut forests \citep{guha2016}.

\section{Population mean depths}
\label{sec:mean}

\begin{theorem}
\label{thm:mean}
Let \(\gamma\) be Euler's constant and let all logarithms be natural. Under the assumptions of Section~\ref{sec:model}, for fixed \(x\in(a,b)\) define
\[
 M_m(x)=2\log m+2\gamma-\tfrac12+\log\bigl(f(x)^2(x-a)(b-x)\bigr).
\]
Then
\begin{equation}
 \mu_m(x)=M_m(x)+o(1),\qquad
 \E D_m^{\mathrm{ins}}(x)=M_m(x)+\tfrac12+o(1).
 \label{eq:means}
\end{equation}
For a fixed held-out query at or beyond a support endpoint \(e\), on its exterior side,
\begin{equation}
 \mu_m(x)=\log m+\gamma+\log\bigl(f(e)(b-a)\bigr)+o(1).
 \label{eq:endpoint}
\end{equation}
For every fixed interior query the right-minus-left bracketing training depth converges in distribution to a nondegenerate law common to these densities.
\end{theorem}

\subsection{One-sided harmonic means}

For uniform data on \([0,1]\), the probability-scale query is denoted by \(p\). Let \(H_k=\sum_{j=1}^k 1/j\) and \(H_0=0\).

\begin{lemma}
\label{lem:harmonic}
For \(k\) iid uniform distances on \([0,\ell]\), their ordered list \(d\) satisfies \(\E\mathcal L(d)=H_k\) and \(\sup_k\E|\mathcal L(d)-H_k|^4<\infty\).
The held-out uniform endpoint mean is exactly \(H_{m-1}\), with bounded fourth moments about \(\log m\).
For a sequence \(p_m\to p\in(0,1)\), the mean of the uniform inserted score at \(p_m\) is
\[
 2\log m+2\gamma+\log(p_m(1-p_m))+o(1).
\]
Its fourth moments about \(2\log m\) are bounded uniformly over queries in any compact interior interval.
\end{lemma}

\begin{proof}
The ordered distances have the joint law \(\ell G_j/G_{k+1}\), where \(G_j\) sums the first \(j\) independent rate-one exponentials. The ratios \(G_{j-1}/G_j\) for \(j\ge2\) have independent \(\Beta(j-1,1)\) laws. At each step beta--gamma independence gives independence of the latest ratio and the current total, and the earlier ratios are independent of that total and the next exponential by induction. The \(j\)th summand has mean \(1/j\), with second and fourth raw moments \(2/[j(j+1)]\) and \(24/[j(j+1)(j+2)(j+3)]\) for \(j\ge2\); the first summand is deterministic. The fourth central moment of the sum is bounded by the sum of the individual fourth central moments plus three times the square of the sum of their variances. Both sums are bounded by the displayed raw-moment series up to absolute factors.

Conditional on the training minimum, the \(m-1\) remaining uniform points are iid above it. The first assertion gives the endpoint mean and moment bound there, and reflection handles the right endpoint. For an inserted query \(p_m\), the two side counts are \(K\) and \(m-K\), with \(K\sim\operatorname{Bin}(m,p_m)\). Conditional side means are \(H_K,H_{m-K}\). For queries in a compact interior interval there exists \(\epsilon>0\) such that \(K,m-K\ge\epsilon m\) except with uniformly exponentially small probability. On this event the harmonic expansion has uniform error \(O(1/m)\), and the log terms \(\log(K/m),\log(1-K/m)\) are bounded. Because \(K/m-p_m\to0\) in probability, their expectations on the event differ from \(\log p_m,\log(1-p_m)\) by \(o(1)\). The complement contributes \(O(m)\) times its probability to the mean and \(O(m^4)\) times its probability to the fourth-moment bound. The same count argument gives this mean expansion for the sum of \(H_{K-1},H_{m-K-1}\) when the counts are positive.
\end{proof}

\subsection{The two bracketing depths}

On the event of two uniform neighbors, let \(A_-=p-z_-\), \(A_+=z_+-p\) and \(g=A_-+A_+\). Let \(V^-\) list the distances from \(z_-\) to the farther points on the left and \(V^+\) the distances from \(z_+\) to the farther points on the right. Define
\begin{equation}
 Q(V,g)=\sum_{j=1}^{|V|}\frac{V_j-V_{j-1}}{V_j}\frac{g}{g+V_j},\qquad V_0=0.
 \label{eq:correction}
\end{equation}
Moving an anchor across the bracket gap changes the outward contribution from \(\mathcal L(V)\) to \(\mathcal L(V)-Q(V,g)\). The gap itself contributes one at either anchor. Consequently the right-minus-left training depth is \(Q(V^+,g)-Q(V^-,g)\), and
\begin{equation}
 D_m(p)=\mathcal L(V^-)+\mathcal L(V^+)+1
        -\frac{A_+}{g}Q(V^+,g)-\frac{A_-}{g}Q(V^-,g).
 \label{eq:held}
\end{equation}

\begin{lemma}
\label{lem:brackets}
For a fixed \(p\in(0,1)\), the uniform held-out depth has mean
\[
 2\log m+2\gamma-\tfrac12+\log(p(1-p))+o(1).
\]
The depth and each bracketing training depth on the two-sided event have bounded fourth moments about \(2\log m\), uniformly on compact interior query intervals. The right-minus-left bracketing depth converges in law to the difference of two nondegenerate corrections which are iid conditional on their common limiting gap.
\end{lemma}

\begin{proof}
Given the counts \(k,l=m-k\), use independent rate-one exponential sequences \(E_i^-,E_i^+\) for the uniform spacings on the two sides. Their totals are \(T_-=\sum_{i=1}^{k+1}E_i^-\), \(T_+=\sum_{i=1}^{l+1}E_i^+\); the scale factors are \(p/T_-\) and \((1-p)/T_+\). The nearest spacings come from \(E_1^-,E_1^+\). The baselines in \eqref{eq:held} have harmonic means \(H_{k-1},H_{l-1}\) and bounded centered fourth moments by the same ratio calculation as in Lemma~\ref{lem:harmonic}, using the exponential sequences after \(E_1^\pm\).

The tail of \eqref{eq:correction} after \(J\) terms is bounded by \(g/V_J\). Indeed, bound each term by \(g(V_j-V_{j-1})/V_j^2\) and sum the integral of \(g/v^2\) from \(V_J\). For \(J=8\), on the count event \(k,l\ge\epsilon m\) and for large enough \(m\), let \(P_J=\sum_{i=2}^{J+1}E_i^-\). Then
\[
 \frac{g}{V_J^-}=\frac{E_1^-}{P_J}
                 +\frac{1-p}{p}\frac{T_-}{P_J}\frac{E_1^+}{T_+}.
\]
The first term has bounded fourth moment by the inverse fourth moment of \(\GammaLaw(8,1)\), with the second parameter denoting rate. For the product in the second term, \(P_J/T_-\sim\Beta(J,k+1-J)\) and \(E_1^+/T_+\sim\Beta(1,l)\) independently. Their inverse and direct fourth moments are \(O(k^4)\) and \(O(l^{-4})\), respectively. Reflection gives the other side's bound. The first \(J\) terms are bounded by \(J\) and each entire correction is at most \(m\) off the count event. Its probability is exponentially small. Equation~\eqref{eq:held} and the corresponding formulas for the two training depths now give the asserted fourth-moment bounds. A missing neighbor off the count event contributes at most \(O(m^4)\) times an exponentially small probability.

For the limit, couple \(K\) across \(m\) by partial sums of iid Bernoulli variables and use the same two exponential sequences for every \(m\). The laws of large numbers give \(K/m\to p\) and \(T_-/K,T_+/(m-K)\to1\) almost surely. Thus \(mA_-,mA_+\) tend to independent rate-one exponentials, and each list \(mV^\pm\) converges term by term to independent rate-one Poisson arrival times, independent of the nearest spacings. The limiting bracket width \(G\) has law \(\GammaLaw(2,1)\). The tail estimate makes the error after \(J\) terms tend to zero almost surely as \(J\to\infty\), since the limiting arrival times diverge. The fourth-moment bound gives uniform integrability for the corrections and their products with the bounded interpolation weights.

If \(\tau_j\) is the \(j\)th outward Poisson arrival, then \(\E[(\tau_j-\tau_{j-1})/\tau_j\mid \tau_j]=1/j\). For independent \(G\sim\GammaLaw(k,1)\) and \(\tau_j\sim\GammaLaw(j,1)\), \(\E[G/(G+\tau_j)]=k/(j+k)\). For \(k=1,2\) the mean correction is therefore
\begin{equation}
 \E Q(\tau,G)=\sum_{j\ge1}\frac{k}{j(j+k)}=H_k.
 \label{eq:gamma}
\end{equation}
For the bracket \(k=2\), the limiting interpolation weights are independent of \(G\) and the outward arrivals, since the ratio of one nearest exponential to their sum is uniform and independent of that sum. The mean offset from the harmonic baselines in \eqref{eq:held} tends to \(1-H_2=-1/2\), proving the expansion by Lemma~\ref{lem:harmonic}. Conditional on \(G\), the two corrections are iid. With every later exponential spacing held fixed, increasing the first outward spacing strictly decreases the first summand in \(Q\) and decreases the remaining summands. Each conditional correction is therefore nondegenerate, as is their difference.
\end{proof}

\subsection{Transfer to physical coordinates}

\begin{lemma}
\label{lem:transfer}
For each side of any training or inserted anchor, the difference between the physical-coordinate side depth and its probability-coordinate counterpart is bounded by a constant depending only on the law and support length. For an iid sequence with dense sample and anchor tending to \(z\), if the side reaches endpoint \(e\ne z\), that difference tends almost surely to
\[
 \log\frac{f(z)|e-z|}{|F(e)-F(z)|}.
 \]
For an interior query, the held-out and both bracketing depths have bounded fourth moments about \(2\log m\), on the two-sided event for the bracketing depths. The held-out endpoint depth has bounded fourth moments about \(\log m\). The inserted fourth-moment bound is uniform on compact interior intervals, and along every deterministic \(x_m\to x\in(a,b)\) its mean is \(M_m(x_m)+\tfrac12+o(1)\).
\end{lemma}

\begin{proof}
On one side of an anchor \(z'\), set \(w(s)=|F(z'\pm s)-F(z')|\) and \(\phi(s)=\log(w(s)/s)\), with \(\phi(0)=\log f(z')\). The formula \(w(s)/s=\int_0^1 f(z'\pm su)\,du\) and \(f_{\min}>0\) give a uniform Lipschitz constant for \(\phi\). In a cell with near and far distances \(0<l<r\), the physical gap fraction minus the probability-scale gap fraction is
\begin{equation}
 \frac{w(l)}{w(r)}\bigl[1-\exp(\phi(r)-\phi(l))\bigr].
 \label{eq:transfer}
\end{equation}
Its absolute value is bounded by a constant times \(r-l\). The first cell on each nonempty side contributes one on either scale. Summing proves the deterministic bound.

Let \(h_{\max}\) be the maximal spacing, including gaps to the support endpoints and the inserted anchor when present. The error from linearizing the exponential in \eqref{eq:transfer} is bounded in sum by a constant times \(\sum (r-l)^2=O(h_{\max})\). In the linear term, replacing \(w(l)/w(r)\) by \(1\) costs \(O(\varepsilon+h_{\max})\) from cells with \(l<\varepsilon\). For the rest, \(1-w(l)/w(r)\le(f_{\max}/f_{\min})h_{\max}/\varepsilon\), giving cost \(O(h_{\max}/\varepsilon)\). The first cell can be included in the telescoping linear sum at cost \(O(h_{\max})\). Since \(h_{\max}\to0\) almost surely for an iid sample with full support, taking \(\varepsilon\downarrow0\) gives \(\phi(0)-\phi(d_{\max})+o(1)\), where \(d_{\max}\) is the farthest distance on the side. As \(z'\to z\) and the farthest point tends to \(e\), this is the claimed limit. The argument applies to deterministic moving anchors as well.

For each bracketing or inserted anchor tending to the same interior \(x\), the sum of the two side corrections tends to
\begin{equation}
 \log\frac{f(x)^2(x-a)(b-x)}{F(x)(1-F(x))}.
 \label{eq:position}
\end{equation}
A bounded correction may be passed through expectation. Both bracketing anchors acquire the same limiting correction, giving convergence of the physical bracket contrast to the same law as on the uniform scale. For interpolation, the physical weight in \eqref{eq:interpolate} differs from the probability-scale weight by at most a constant times the bracket width. This follows by writing the probability-scale weight as the physical subinterval fraction times the ratio of two averages of \(f\) on subintervals of the bracket. The weight error tends to zero almost surely, is bounded, and multiplies a contrast with bounded second moment by Lemma~\ref{lem:brackets}. Cauchy--Schwarz shows the interpolation error tends to zero in expectation. For fourth moments use the deterministic bound on each anchor and convexity for the interpolation with physical weights. The exceptional one-sided event for an interior query has exponentially small probability and depths bounded by \(m\).

For an endpoint use the training extreme as anchor, so there is one side and the coordinate limit is \(\log(f(e)(b-a))\). For a moving inserted query \(x_m\to x\), apply Lemma~\ref{lem:harmonic} with \(p_m=F(x_m)\). On the physical scale the deterministic bound and the dense-sample limit give expectation correction \eqref{eq:position} at \(x\). The same correction evaluated at \(x_m\) differs by \(o(1)\) by continuity. These bounds apply uniformly to the fourth moments over any compact interior interval, because both the coordinate bound and the harmonic count estimate are uniform there.
\end{proof}

\begin{proof}[Proof of Theorem~\ref{thm:mean}]
Lemmas~\ref{lem:harmonic} and \ref{lem:brackets} give the uniform means and the bracket contrast. Lemma~\ref{lem:transfer} passes them to physical coordinates. The probability-scale position term cancels in the interior. At or beyond an endpoint the clamping identity gives the same depth at every query on the corresponding exterior ray.
\end{proof}

\subsection{Comparison with Zheng et al.}
\label{sec:published}

Theorem 4.1 of \citet{zheng2026} gives the same density and position terms for a continuous distribution on a single bounded interval, and states sample-size and additive terms \(2\log m+2+2\gamma\) in the interior and \(\log m+1+\gamma\) elsewhere. For densities bounded away from zero and Lipschitz on the interval, our formulas decrease these constants by \(5/2\) and \(1\).

In Appendix B.1 of \citet{zheng2026}, equations (3)--(4) approximate the endpoint gap sum by an integral starting at the first sample spacing. For a uniform sample, conditional on the extremes, that spacing divided by the sample range has law \(\Beta(1,m-2)\) for \(m\ge3\). The expected integral is \(H_{m-2}\). With the first gap's unit contribution, the approximation gives \(1+H_{m-2}\), while Lemma~\ref{lem:harmonic} gives the exact \(H_{m-1}\). The discrepancy tends to one because the near-anchor denominators shrink with the mesh. For the interior, Appendix B.1 uses Lemma B.2 to discard the interpolation. That lemma bounds a reciprocal-square sum by an integral singular at the anchor. The nondegenerate limit in Theorem~\ref{thm:mean} contradicts the asserted almost-sure vanishing of the maximal adjacent contrast, since that maximum dominates the absolute contrast at any fixed interior query. The cancellation of the common interior constant preserves the mean-difference formula of their Corollary 4.2.

Supplement Table S1 compares the finite-sample means to both formulas. Its two interior designs suggest residuals of order \(1/m\), and its nonuniform endpoint is compatible with a \((\log m)/m\) correction. At the uniform endpoint the harmonic formula gives the exact leading residual \(-1/(2m)\).

\begin{corollary}[Two separated intervals]
\label{cor:two}
Let the support be \([a,b]\cup[c,d]\) with \(a<b<c<d\) and a positive Lipschitz density on each component. For a component \(I=[l,r]\), put
\[
 R_{[a,b]}(z)=\frac{c-b}{c-z}+\log\frac{d-z}{c-z},\qquad
 R_{[c,d]}(z)=\frac{c-b}{z-b}+\log\frac{z-a}{z-b}.
\]
For \(x\in(l,r)\), the held-out mean equals
\[
 2\log m+2\gamma-\tfrac12+\log\bigl(f(x)^2(x-l)(r-x)\bigr)+R_I(x)+o(1).
\]
At an endpoint \(e\) of \(I\), its asymptotic expression is
\[
 \log m+\gamma+\log\bigl(f(e)(r-l)\bigr)+R_I(e).
\]
Between \(b\) and \(c\) the asymptotic expression interpolates linearly between the two endpoint expressions, and outside \([a,d]\) it clamps to the corresponding outer endpoint.
\end{corollary}

\begin{proof}
Condition on the counts in the two components. For \(I\) of mass \(\pi_I>0\) the count is binomial with fraction converging to \(\pi_I\), and the conditional density is \(f/\pi_I\). Apply the one-component formulas to the interior query or the component's relevant training extreme. The random count and conditional density combine to give \(\log m+\log f(e)\) at an edge and \(2\log m+2\log f(x)\) in the interior. Indeed the count divided by \(m\) is bounded away from zero with exponentially high probability, so the log converges in expectation on that event and the deterministic depth bound controls its complement. On the same count event the centered extreme depths have bounded fourth moments by Lemma~\ref{lem:transfer}.

For an anchor tending to \(z\) in the left component, let \(u\) be the last point there and \(v\) the first in the right component. The single gap \((u,v)\) contributes \((v-u)/(v-z')\) for anchor \(z'\), tending to \((c-b)/(c-z)\). Beyond \(v\), the sum is a right-endpoint Riemann sum with denominator bounded below by \(c-b\). It tends to \(\int_c^d (t-z)^{-1}dt\). The added terms are bounded by the full range divided by \(c-b\), which permits taking expectations and preserves the extreme-depth moment bounds. Reflection gives the right-component expression. When the query is in a component interior its bracketing anchors lie in that component except on an exponentially small event. For any fixed \(x\in[b,c]\), the same \(u,v\) bracket the query and their interpolation weight tends to \((c-x)/(c-b)\). The two depths have the common leading term \(\log m\) and bounded centered second moments, so the replacement of the weights holds in expectation by Cauchy--Schwarz. Outside the total support clamping always applies, and an empty component has exponentially small probability.
\end{proof}

In Corollary 4.3 of \citet{zheng2026}, the two component interiors thus require both the shift \(-5/2\) and the separate gap fraction in \(R_I\). The outer edges require the shift \(-1\) and their corresponding gap fraction. At \(b,c\) this fraction is one, canceling the one-sided shift of one cut. The correct gap-adjacent logarithmic terms are \(\log((b-a)(d-b)/(c-b))\) and \(\log((d-c)(c-a)/(c-b))\), respectively. The printed expression at \(b\) uses \(d-c\) in the denominator.

\section{Variance across reused reference pools}
\label{sec:variance}

For this section the support is a single interval \([a,b]\). Let \(\mathcal X_N=(X_1,\ldots,X_N)\) be an iid reference pool. Define the complete subbagging score
\begin{equation}
 U_{N,m}(x)=\binom Nm^{-1}
           \sum_{\substack{I\subseteq\{1,\ldots,N\}\\ |I|=m}}
            D(x;\{X_i:i\in I\}).
 \label{eq:u}
\end{equation}
This \(U\)-statistic has mean \(\mu_m(x)\), and its variance is over fresh pools. If two size-\(m\) subsets are drawn independently conditional on the pool, the covariance of their cut-averaged depths equals \(\Var(U_{N,m}(x))\), since the draws have common conditional mean \(U_{N,m}(x)\). For an average of \(J_{\rm tree}\) conditionally independent trees with both subset and cut randomness, the additional variance is \(\E[\Var(Y_{\rm tree}\mid\mathcal X_N)]/J_{\rm tree}\), by total variance for a single tree depth \(Y_{\rm tree}\). Here we study the pool term after taking both averages. Proposition 3.1 and Section 6.1 of \citet{zheng2026} consider tree averaging on a fixed dataset.

\subsection{Rates and fixed-query comparisons}

Let \(q(x)=\min(b,\max(a,x))\) be the effective location, since all queries on a given exterior ray have the same depth as its endpoint. Define \(\kappa_q=\kappa_{\mathrm{int}}\) for \(a<q<b\) and \(\kappa_q=\kappa_{\partial}\) at an endpoint. The two positive constants are given by \eqref{eq:constants} below.

\begin{theorem}
\label{thm:variance}
Under the assumptions of Theorem~\ref{thm:mean}, fix a query \(x\). For all \(N\ge m\) and sufficiently large \(m\), \(\Var(U_{N,m}(x))\asymp m/N\), with positive comparison constants independent of \(m,N\). If \(x'\) and \(x''\) have distinct effective locations, then
\begin{equation}
 \Var\bigl(U_{N,m}(x')-U_{N,m}(x'')\bigr)\asymp m/N.
 \label{eq:rankrate}
\end{equation}
Along \(m\to\infty\) and \(m/N\to0\), the ratios of these variances to \(m/N\) converge to \(\kappa_{q(x)}\) and \(\kappa_{q(x')}+\kappa_{q(x'')}\), respectively. In this regime the covariance of the two distinct effective query scores is \(o(m/N)\). The variance constants are independent of \(f\) and of the fixed location within each query type.

For either statistic, convergence in probability to its mean along \(m\to\infty\) holds exactly when \(m/N\to0\). The same criterion holds with the corresponding asymptotic mean expression from Theorem~\ref{thm:mean}, or its difference for two queries.
\end{theorem}

At \(m=N=256\), the two-query example in Supplement Table S2 has contrast variance \(1.079\) cuts squared, so its pool standard deviation is \(1.04\) cuts, comparable to the one-cut density contribution \(2\log(e^{1/2})=1\) from a density ratio \(e^{1/2}\approx1.65\) at equal centrality.

For two fixed interior queries the population depth difference converges to the difference of their log-density and log-centrality terms. When this limit is nonzero, \(m/N\to0\) suffices for the probability of a correct ordering to tend to one. If \(m/N\) stays bounded away from zero, the random contrast around the population contrast has variance bounded away from zero. This describes the scale of ranking uncertainty and gives no reversal probability for an arbitrary fixed nonzero gap. For the score \(2^{-U_{N,m}(x)/\nu_m}\) with the common positive normalizer \(\nu_m\sim 2\log m\) in \citet{liu2008,liu2012}, the centered normalized depth converges to zero in mean square by Theorem~\ref{thm:variance}. For every \(m\) the transformation is strictly decreasing, so ordering by increasing depth and by decreasing score gives the same pairwise comparison. The result concerns trees grown to singletons and fixed effective locations.

\subsection{Local effect of forcing one observation}

Let \(S\) contain \(m-1\) iid observations. For \(q\in[a,b]\) and \(y\in[a,b]\) put
\begin{equation}
 \Delta_{m,q}(y)=D(q;S\cup\{y\})-D(q;S),\qquad
 \psi_{m,q}(y)=\E\Delta_{m,q}(y).
 \label{eq:force}
\end{equation}

\begin{lemma}
\label{lem:force}
If \(q\in(a,b)\), then \(\Delta_{m,q}(y)\) has uniformly bounded fourth moments for \(y\) in a fixed compact neighborhood of \(q\) inside the support. For an endpoint \(q\), the fourth moments are bounded uniformly in \(y\in[a,b]\). For each effective location \(q\) there is a constant \(C_q\) such that, for all \(m\ge2\) and all \(y\in[a,b]\),
\begin{equation}
 |\psi_{m,q}(y)|\le \frac{C_q}{1+m|y-q|}.
 \label{eq:envelope}
\end{equation}
\end{lemma}

\begin{proof}
For any old training anchor, insertion within a gap whose near and far distances are \(l<r\) changes its depth by
\[
 (h-l)(1/h-1/r),\qquad l<h<r,
\]
where \(h\) is the distance to the inserted point. This is in \([0,1]\); if insertion extends the range, use \(1/r=0\). Thus on the event that \(S\) brackets an interior \(q\), the centered fourth moments of all potential new bracketing anchors are bounded. For the old anchors this follows from Lemma~\ref{lem:transfer} and the deterministic increment bound. For the new anchor \(y\), use the inserted fourth-moment bound on the compact neighborhood in that lemma. Both the original and forced query depths are convex combinations of the corresponding anchor depths, or the inserted depth itself when \(y=q\). Their difference has bounded fourth moment by the triangle inequality in \(L^4\), with centering at \(2\log m\), since \(2\log(m-1)-2\log m\) is bounded. This also bounds the absolute contrast between any two eligible bracketing depths in the augmented tree. The one-sided event has exponentially small probability and depth bounded by \(m\).

At \(q=a\) the score is the depth of the sample minimum \(Z\). If \(y>Z\), the old anchor increment is bounded by one. If \(y<Z\), the new minimum has increment \(1-Q(V,Z-y)\), where \(V\) lists the distances from \(Z\) to the remaining old points. For \(J=8\) the tail bound gives \(Q(V,Z-y)\le J+(Z-a)/V_J\) if there are at least \(J\) terms, and \(Q(V,Z-y)\le J\) otherwise. On transforming to the uniform scale, the ratio is bounded by \(f_{\max}/f_{\min}\) times the corresponding uniform ratio. With normalized exponential spacings, that ratio is \(E_1/(E_2+\cdots+E_{J+1})\). Its fourth moment is finite. Reflection gives the same conclusion at \(b\).

For the envelope, let \(s=|y-q|>0\) and first take \(y>q\). In the interior, with probability at least \(1-2e^{-\beta_q ms}\), an old training point on each side lies within \(s/2\) of \(q\), for some \(\beta_q>0\). Indeed the intervals truncated to the support each have length at least a fixed positive multiple of \(s\), since \(s\le b-a\) and \(q\) is fixed. At the left endpoint use the event \(Z\le a+s/2\), with the same exponential bound. On these events the anchors and weights stay the same after insertion, and each anchor lies at least \(s/2\) to the left of \(y\). The increment at each anchor is bounded by \(2B_y/s\), where \(B_y\) is the distance back to the closest old observation to the left of \(y\), using \(a\) if none exists. This follows from the increment formula and \(1/h-1/r\le 1/h\); the preceding point is at or beyond each anchor. The tail of \(B_y\) is bounded by \(e^{-(m-1)f_{\min}u}\) over \(0<u<y-a\), so \(\E B_y\le [(m-1)f_{\min}]^{-1}\).

On the complementary event use Cauchy--Schwarz and the second moments just established when \(y\) is near \(q\), and for all \(y\) at the endpoint. For \(y\) outside the interior neighborhood, \(s\) is bounded below by a positive number, so the crude bound \(m\) times \(2e^{-\beta_q ms}\) is also exponentially small in \(ms\), with a modified exponent. Reflection treats insertions to the left. Combining the bounds gives \(O((ms)^{-1})+O(e^{-\beta'_q ms})\) when \(ms\ge1\). For \(ms<1\) the local moment bound suffices, enlarging the constant for finitely many small \(m\) if needed.
\end{proof}

\subsection{The limiting Poisson profiles}

For a unit-rate Poisson process on the line, truncate it to a bounded window around zero, with depth zero for an empty window. Score the query zero before and after adding a point at \(t\). The limiting expected depth difference as the window expands is denoted by \(\Psi_{\mathrm{int}}(t)\). For an endpoint, use a unit-rate process on the positive half-line and \(t\ge0\), with the query clamped to the first observation, to obtain \(\Psi_{\partial}(t)\). Then
\begin{equation}
 \kappa_{\mathrm{int}}=\int_{\mathbb R}\Psi_{\mathrm{int}}(t)^2\,dt,\qquad
 \kappa_{\partial}=\int_0^\infty\Psi_{\partial}(t)^2\,dt .
 \label{eq:constants}
\end{equation}

\begin{lemma}[Local Poisson limit]
\label{lem:local-limit}
For fixed \(q\) in the interior and \(t\in\mathbb R\), the limit defining \(\Psi_{\mathrm{int}}(t)\) exists and
\[
 \psi_{m,q}\bigl(q+t/(m f(q))\bigr)\longrightarrow \Psi_{\mathrm{int}}(t).
\]
At \(q=a\) the same convergence holds to \(\Psi_\partial(t)\) for \(t\ge0\); at \(q=b\) use \(y=b-t/(m f(b))\).
\end{lemma}

\begin{proof}
On the probability scale the normalized exponential coupling of Lemma~\ref{lem:brackets}, used with \(m-1\) points, gives almost-sure convergence of every fixed number of first arrivals on each side. The left and right limits are independent unit-rate Poisson processes. At the left endpoint use the single normalized exponential sequence starting at zero, and reflect at the right endpoint. The inverse distribution function gives the same limits for physical distances scaled by \(m f(q)\), since \(F(q+s)-F(q)=f(q)s+o(s)\). In a fixed window the ordered configurations converge almost surely: the first limiting arrival outside each boundary rules out extra points, and a Poisson arrival lies on the boundary with probability zero. The truncated depth differences are continuous at these configurations almost surely, since the gaps are strictly positive and the forced point does not coincide with a Poisson observation.

To bound the terms discarded by truncation, fix \(M>|t|\) and suppose the original scoring anchors are all in \([-M,M]\). After insertion the scoring anchors stay in that interval. Take \(L>2M\) and suppose there is a retained observation between \(L/2\) and \(L\) on each relevant side. A discarded gap on the right, including a gap crossing the truncation boundary, then has endpoints \(u,v\) with \(v>u\ge L/2\). The contribution at an anchor \(z\in[-M,M]\) differs from its value with denominator \(v\) by at most
\[
 \left|\frac{v-u}{v-z}-\frac{v-u}{v}\right|
 \le \frac{M(v-u)}{(v-M)^2}.
\]
Summing over discarded gaps bounds the difference on that side by \(M\int_{L/2}^{\infty}(s-M)^{-2}\,ds=M/(L/2-M)\). On the left apply the same estimate after reflection. The discarded gaps are the same before and after insertion. Their reference contributions, with denominators given by the far-end distances to zero, cancel when subtracting the two depths, since the interpolation weights for each depth sum to one. Thus the error in a paired depth difference is at most \(4M/(L/2-M)\) for two sides, also valid on a half-line.

This bound applies to two nested truncations and to the full finite sample and its truncation. Almost surely a finite \(M\) contains the original Poisson anchors, and all sufficiently large dyadic radii have observations in the required bands. For each such fixed radius the coupled finite configurations eventually retain their original anchors and have observations in the same bands. The bound also controls every larger radius between the dyadic radii. Taking the finite-sample limit in each fixed window and then increasing \(L\) proves convergence of the full scaled depth difference to the limit of the Poisson truncations. Lemma~\ref{lem:force} gives uniform integrability for the finite-sample depth differences.

For the Poisson expectations, once the original anchors are present, an insertion outside the interior bracket has a depth increment of at most one by the gap update. An insertion inside the bracket leaves the two outward baseline sums common to the old and new anchor depths. There are at most two unit contributions from gaps inside the bracket and at most two outward corrections of the form \(Q\), each with shift no greater than the old width \(G\). Thus the paired difference is dominated by \(4+2Q(\tau^-,G)+2Q(\tau^+,G)\), where \(\tau^\pm\) are the full outward arrival sequences. This bound is integrable by \eqref{eq:gamma}. On the half-line an insertion before the old minimum \(Z\) is bounded by \(1+Q(\tau,Z)\), also integrable by \eqref{eq:gamma}. If a window omits an original anchor, its contribution to the expected absolute difference tends to zero by Cauchy--Schwarz: the missing-anchor probability is exponentially small in \(L\) and the Poisson count plus one has second moment \(O((1+L)^2)\). These bounds justify taking the Poisson expectation in the limit.
\end{proof}

\begin{lemma}[Explicit profiles]
\label{lem:profiles}
For \(s>0\), define
\begin{equation}
 \chi(s)=\int_0^\infty\frac{1-e^{-v}}{v}\frac{s}{s+v}\,dv,\qquad
 \rho(s)=\frac1s-\E\frac1{s+E},\quad E\sim\operatorname{Exp}(1),
 \label{eq:kr}
\end{equation}
and set \(\chi(0)=0\). The endpoint profile is
\begin{equation}
 \Psi_\partial(t)=\int_0^t e^{-z}(1-e^{-(t-z)})\rho(t-z)\,dz .
 \label{eq:boundaryprofile}
\end{equation}
The interior profile is even. For \(t\ge0\) let \(A_-,A_+\) be independent rate-one exponentials and \(g=A_-+A_+\). Its value is the expectation of
\begin{equation}
 (1-e^{-s})\bigl[\lambda\rho(t+A_-)+(1-\lambda)\rho(s)\bigr],
 \quad s=t-A_+,\quad \lambda=A_+/g ,
 \label{eq:profileout}
\end{equation}
on \(t>A_+\), and of
\begin{equation}
 \alpha\,\frac{g_2}{g}
   +(1-\alpha)\bigl[1+\chi(g)-\chi(g_1)-\chi(g_2)\bigr],
 \quad g_1=A_-+t,\quad g_2=A_+-t,\quad \alpha=t/g_1 ,
 \label{eq:profilein}
\end{equation}
on \(t<A_+\), with each contribution weighted by its event probability. The constants in \eqref{eq:constants} are finite and strictly positive.
\end{lemma}

\begin{proof}
For the rate-one outward arrivals \(\tau_j\), the \(j\)th gap fraction conditional on \(\tau_j\) has mean \(1/j\). Summing the gamma densities with weights \(1/j\) gives \((1-e^{-v})/v\), so \(\chi(s)=\E Q(\tau,s)\). For numerical evaluation one can use
\[
 \chi(s)=\gamma+\log s+e^s\int_s^\infty e^{-v}\,\frac{dv}{v}.
\]
Both expressions vanish at zero. Differentiating the integral and integrating by parts gives the common derivative
\[
 \int_0^\infty \frac{1-e^{-v}}{(s+v)^2}\,dv
 = e^s\int_s^\infty e^{-v}\,\frac{dv}{v}.
\]

At an endpoint the minimum \(Z\) is exponential. Given \(Z>t\), the shift \(Z-t\) is rate-one exponential and independent of the process beyond \(Z\), so the mean increment is \(1-H_1=0\) by \eqref{eq:gamma}. If \(Z<t\), the backward gap from \(t\) is the smaller of \(t-Z\) and a rate-one exponential. Its mean is \(1-e^{-(t-Z)}\), and the independent forward gap is exponential. The gap update at \(Z\) has conditional mean \((1-e^{-(t-Z)})\rho(t-Z)\), giving \eqref{eq:boundaryprofile}.

In the interior, reflection gives the claimed evenness. Given the two original neighbors, insertion at \(t>A_+\) leaves both interpolation anchors unchanged. The backward gap has mean \(1-e^{-s}\) by the same truncated exponential calculation with \(A_+\) as the leftmost point to the right of zero. The independent forward gap is exponential, so weighting the two anchor increments gives \eqref{eq:profileout}. If \(t<A_+\), the left anchor acquires \(g_2/g\) by splitting the bracket gap, and the new anchor at \(t\) has two first gaps of contribution one each. The outward baseline sums stay the same. The expected outward correction at the new anchor is \(\chi(g_1)+\chi(g_2)\) in place of the old query's expected correction \(\chi(g)\). The new weight on the left anchor is \(\alpha\), giving \eqref{eq:profilein}.

At zero, \eqref{eq:gamma} gives \(\Psi_{\mathrm{int}}(0)=1+H_2-2H_1=1/2\). As \(t\downarrow0\), the term \eqref{eq:profileout} is bounded by one on an event of vanishing probability, and \eqref{eq:profilein} converges by domination by \(2+3\chi(g)\). Thus \(\Psi_{\mathrm{int}}\) is continuous at zero. The endpoint profile is strictly positive for \(t>0\) because \(\rho(s)>0\) for \(s>0\). Lemma~\ref{lem:local-limit} transfers the square-integrable envelope \eqref{eq:envelope} to both profiles. This proves positivity and finiteness of both constants.
\end{proof}

\begin{proof}[Proof of Theorem~\ref{thm:variance}]
The first projection of the kernel at effective location \(q\) is
\[
 \eta_{m,q}(y)=\E D(q;S\cup\{y\})-\mu_m(q)
             =\psi_{m,q}(y)-\E\psi_{m,q}(X_1).
\]
The envelope \eqref{eq:envelope} gives \(\E|\psi_{m,q}(X_1)|=O(\log m/m)\). In \(m\E\psi_{m,q}(X_1)^2\), change variables by writing \(y=q+t/(m f(q))\), reflected at \(b\). The resulting density factor \(f(y)/f(q)\) tends to one and is bounded. Lemma~\ref{lem:local-limit} and the square of \eqref{eq:envelope} give dominated convergence. The squared centering term times \(m\) vanishes, so \(m\E\eta_{m,q}(X_1)^2\to\kappa_q\). For distinct \(q,q'\), split at the midpoint between them. On either piece at least one profile is \(O(1/m)\) uniformly. The \(L^1\) bounds and the centering terms therefore give
\[
 m\E\!\left[\eta_{m,q}(X_1)\eta_{m,q'}(X_1)\right]\longrightarrow 0.
\]
The leading kernel projection for a contrast has variance constant \(\kappa_q+\kappa_{q'}\).

For either kernel \(h_m\) (one query or the contrast), let \(v_{m,j}\) denote the variance of its order-\(j\) canonical Hoeffding kernel \citep{hoeffding1948}. Its full variance is \(\sum_{j=1}^m\binom mj v_{m,j}\), since the components are orthogonal by conditional centering in every argument. Counting how often a fixed \(j\)-subset occurs in \eqref{eq:u} gives its coefficient \(\binom mj/\binom Nj\) and the variance formula
\begin{equation}
 \Var(U)=\frac{m^2}{N}v_{m,1}+
 \sum_{j=2}^m \frac{\binom mj}{\binom Nj}\binom mj v_{m,j}.
 \label{eq:hoeffding}
\end{equation}
Here \(m v_{m,1}\) tends to the positive constant just proved. Lemma~\ref{lem:transfer} bounds \(\Var(h_m)\) uniformly, subtracting \(2\log m\) or \(\log m\) at each effective location. The binomial ratio is at most \((m/N)^j\), so the variance of the terms of order at least two is at most \((m/N)^2\Var(h_m)\). These two facts give the upper and lower rates and the small-fraction limits. The covariance limit for distinct locations follows by the variance identity for the contrast.

Conditional Jensen bounds the fourth moment of the centered statistic by the fourth moment of its centered kernel. These moments are bounded by Lemma~\ref{lem:transfer}, also for the contrast. If \(m/N\to0\), convergence to the mean follows from Chebyshev's inequality. Otherwise a subsequence has variance bounded away from zero. Convergence in probability of the centered statistic to zero on that subsequence would force its variance to vanish by uniform integrability of the squares. Theorem~\ref{thm:mean} also gives vanishing errors in the asymptotic mean expressions.
\end{proof}

\section{Discussion and limitations}

Our results concern one-dimensional isolation trees grown to singleton leaves and evaluated at fixed queries. Standard iForest caps the tree height at \(\lceil\log_2\psi\rceil\) for subsample size \(\psi\) and adds \(c(n_{\mathrm{leaf}})\) to the traversed depth at a terminal node containing \(n_{\mathrm{leaf}}\) observations \citep[Algorithms 1--3]{liu2008}. Early stopping and the terminal adjustment change the depth kernel, so whether the mean corrections and variance rates persist for the adjusted path length is open. In multiple dimensions, a split on one feature changes the sample and spacings available for subsequent splits on other features. Extending the gap analysis to these trees, and measuring how finite forest size and pool reuse jointly affect rankings, are directions for further work.

\section*{Code and data availability}

The simulation and quadrature code, run instructions, and checksums for the numerical reports are available at \url{https://github.com/sehejjain/isolation-tree-bagging}. All data are synthetic and generated by the code. The numerical results and Monte Carlo uncertainties appear in the supplement.

\bibliography{references}
\bibliographystyle{tmlr}

\clearpage
\raggedbottom
\appendix
\renewcommand{\thetable}{S\arabic{table}}

\section{Convergence of the mean depths}

Write \(M_m(x)\) for the corrected held-out mean asymptote in Theorem 1, including its logarithmic growth, density and position terms. Table~\ref{tab:convergence} reports the residual from that asymptote for uniform and nonuniform designs on \([0,1]\). The interior queries use the same probability coordinate \(p=0.3\). They are \(x=0.3\) for the uniform law and \(x=\sqrt{0.85}-0.5\approx0.422\) for density \(f(t)=\tfrac12+t\). Both left-endpoint queries are \(x=0\).

\begin{table}[htbp]
\centering
\small
\caption{Estimated mean minus the corrected asymptote, in cuts. Parentheses give Monte Carlo standard errors. The last row shows the published asymptote on this same scale. The uniform endpoint entries are exact up to rounding.}
\label{tab:convergence}
\setlength{\tabcolsep}{5pt}
\begin{tabular}{@{}lrrrr@{}}
\toprule
 & \multicolumn{2}{c}{Interior queries} & \multicolumn{2}{c}{Left endpoints}\\
\cmidrule(lr){2-3}\cmidrule(l){4-5}
\(m\) & Uniform & \(f(t)=\frac12+t\) & Uniform & \(f(t)=\frac12+t\)\\
\midrule
32   & \(0.02372\;(0.00075)\) & \(-0.01736\;(0.00076)\) & \(-0.01571\) & \(0.23591\;(0.00026)\)\\
64   & \(0.01384\;(0.00040)\) & \(-0.01315\;(0.00041)\) & \(-0.00783\) & \(0.15636\;(0.00018)\)\\
128  & \(0.00734\;(0.00023)\) & \(-0.00810\;(0.00024)\) & \(-0.00391\) & \(0.09837\;(0.00011)\)\\
256  & \(0.00357\;(0.00015)\) & \(-0.00488\;(0.00015)\) & \(-0.00195\) & \(0.05978\;(0.00007)\)\\
1024 & \(0.00100\;(0.00007)\) & \(-0.00125\;(0.00007)\) & \(-0.00049\) & \(0.02024\;(0.00002)\)\\
4096 & \(0.00027\;(0.00003)\) & \(-0.00030\;(0.00003)\) & \(-0.00012\) & \(0.00641\;(0.00001)\)\\
\midrule
Published & \(2.5\) & \(2.5\) & \(1\) & \(1\)\\
\bottomrule
\end{tabular}
\end{table}

Across \(m=256\) to \(4096\), the effective exponents \(-\log(|r_{4096}/r_{256}|)/\log(16)\) for the absolute residuals are approximately \(0.93\) in the uniform interior, \(1.01\) in the nonuniform interior, and \(0.81\) at the nonuniform endpoint. At that nonuniform endpoint the product \(m r_m\) increases from \(15.3\) to \(26.3\) over this range, consistent with a logarithmic factor. These are finite-range empirical rates for the listed designs. At the uniform endpoint the harmonic expansion gives the exact leading error, \(H_{m-1}-\log m-\gamma=-1/(2m)+O(m^{-2})\).

For each size we use \(32{,}768\) iid replicates and average over cuts exactly. To resolve the smaller biases we compute the conditional harmonic baselines exactly and use paired differences for the gap corrections. In each two-sided uniform sample the finite and Poisson corrections share the exponential spacings before normalization. For the Poisson control we evaluate its first 24 outward terms and integrate the remaining tail. Given the last included outward coordinate \(w\) and bracket width \(g\), the expected tail is \(\chi(w+g)-\chi(w)\), using \(\chi\) from the main paper. Indeed the future arrival at distance \(v\) from \(w\) has expected increment \(v/j\) conditional on being the \(j\)th future arrival. Summing those densities yields the integrand \(g(1-e^{-v})/((w+v)(w+g+v))\). The unconditional weighted Poisson correction is \(3/2\), independent of the two side counts used by the finite sample. For a one-sided uniform sample we use its exact conditional harmonic mean. For the nonuniform entries, we add the paired depth change under the physical quantile map, using the same samples as the corresponding uniform control and the same probability coordinate as the uniform query.

A separate simulation with support \([0,1]\cup[4,6]\), weights \(0.4,0.6\) and uniform component densities compares the two-support formula with cut-averaged means from \(32{,}768\) samples of size \(2048\). The seven query locations are \(0,0.3,1,2,4,5,6\), including both bracket endpoints and the gap interior. All seven estimates differ from the asymptotes by fewer than \(0.0044\) cuts. Their Monte Carlo standard errors range from \(0.0025\) to \(0.0041\) cuts.

We repeat the check with two nonuniform component densities, \(0.4(\tfrac12+x)\) on \([0,1]\) and \(0.3(1.5-(x-4)/2)\) on \([4,6]\). They have the same component masses and opposite slopes. Using \(32{,}768\) samples of size \(4096\), the residual of largest magnitude over the same seven queries is \(+0.0130\) cuts at \(x=0\). Its Monte Carlo standard error is \(0.0033\) cuts, so this residual is about four standard errors from zero. Conditional on membership in the left component, the density is the same as in the nonuniform endpoint design of Table~\ref{tab:convergence}, with an expected component count of \(0.4m\). Rescaling the single-interval residual at \(m=4096\) to that count using the suggested \((\log m)/m\) behavior gives about \(0.014\) cuts, a plausible scale for the finite-sample bias. Across the seven queries, the Monte Carlo standard errors range from \(0.0025\) to \(0.0041\) cuts. At the gap query \(x=2\), the residual is \(-0.0027\) cuts with standard error \(0.0025\).

\section{Variance constants and shared-pool checks}

Numerical quadrature of the squared Poisson profiles gives \(\kappa_{\mathrm{int}}\approx0.4548\) and \(\kappa_{\partial}\approx0.1526\). The corresponding constant for a contrast between two distinct interior queries is about \(0.9096\). The integrals use Gauss--Laguerre rules of orders 96 and 192 for the exponential expectations and adaptive quadrature for the remaining integrals. Exponentially weighted regions beyond an exponential coordinate of 40 are omitted. The two evaluations of \(\kappa_{\mathrm{int}}\) differ by about \(1.5\times10^{-5}\). Direct paired Monte Carlo estimates of each profile at \(t=0,0.5,2,4\), using the first 128 Poisson arrivals on each relevant side, check the conditional-increment formulas numerically. The eight estimates are within \(0.009\) cuts of the limiting profiles, with Monte Carlo standard errors from \(0.0004\) to \(0.0038\) cuts. A finite number of arrivals also introduces truncation error.

\begin{table}[htbp]
\centering
\small
\caption{Shared-pool estimates of \((N/m)\Var(U)\) for \(f(t)=\frac12+t\), \(m=256\). Parentheses give Monte Carlo standard errors. For \(N=m\) each statistic is its original single-bag kernel, averaged over cuts.}
\label{tab:variance}
\begin{tabular}{@{}rrrr@{}}
\toprule
\(N\) & Interior \(x=0.3\) & Endpoint \(x=0\) & Contrast \(0.3\) vs.\ \(0.7\)\\
\midrule
256  & \(0.5237\;(0.0041)\) & \(0.3509\;(0.0027)\) & \(1.0790\;(0.0085)\)\\
1024 & \(0.4416\;(0.0122)\) & \(0.1547\;(0.0078)\) & \(0.8876\;(0.0245)\)\\
4096 & \(0.4103\;(0.0476)\) & \(0.1768\;(0.0307)\) & \(0.8607\;(0.0958)\)\\
\midrule
Small-fraction limit & \(0.4548\) & \(0.1526\) & \(0.9096\)\\
\bottomrule
\end{tabular}
\end{table}

For each row of Table~\ref{tab:variance}, we generate \(32{,}768\) independent pairs of bags using the exact hypergeometric distribution of their overlap. Conditional on that count, the overlap and the two sets of exclusive observations are independent draws. Within each bag the two queries used in the contrast share the same observations. The sample covariance of the two bag scores estimates the complete subbagging variance, by conditioning on the common pool. The \(N/m\) scaling also multiplies the Monte Carlo errors.

\subsection{Direct finite-bag projection}

The shared-pool covariance in Table~\ref{tab:variance} has a standard error of about \(0.05\) for the interior query at \(N=4096\). To check the leading constants more precisely, we evaluate the finite-sample influence \(\psi_{m,q}\) directly by paired insertion for \(f(x)=\tfrac12+x\) on \([0,1]\), at \(q=0.3\) and \(q=0\). At each forced location \(y\), the depths with and without the forced point share the same \(m-1\) observations. The integration coordinate \(t=m(F(y)-F(q))\) gives \(m\,dF(y)=dt\), so the raw target is \(m\E[\psi_{m,q}(X)^2]\). The centered target \(m\Var(\psi_{m,q}(X))\) is the scaled first-projection variance in Theorem 3.

We use two independent groups of \(1024\) bags per block and share each group's bags over the grid. The product of the two mean profiles estimates \(\psi_{m,q}^2\) at each location without the bias of squaring a noisy mean. For the centered estimate, subtract the product of the two integrated mean profiles divided by \(m\). We average over 48 independent blocks and compute standard errors across blocks. Composite Simpson quadrature covers the entire support in the \(t\) coordinate, using spacing \(1/8\) out to \(|t|=8\) and eight subintervals per dyadic band in the tails. Comparing to the grid with every other point removed checks sensitivity to spacing.

\begin{table}[htbp]
\centering
\small
\caption{Direct finite-bag estimates of \(m\E[\psi_{m,q}(X)^2]\) (raw) and \(m\Var(\psi_{m,q}(X))\) (centered). Parentheses give Monte Carlo standard errors.}
\label{tab:projection}
\begin{tabular}{@{}lrrrr@{}}
\toprule
 & \multicolumn{2}{c}{Interior \(q=0.3\)} & \multicolumn{2}{c}{Endpoint \(q=0\)}\\
\cmidrule(lr){2-3}\cmidrule(l){4-5}
\(m\) & Raw & Centered & Raw & Centered\\
\midrule
256 & \(0.4566\;(0.0014)\) & \(0.4407\;(0.0014)\) & \(0.1492\;(0.0014)\) & \(0.1455\;(0.0014)\)\\
1024 & \(0.4562\;(0.0016)\) & \(0.4522\;(0.0016)\) & \(0.1525\;(0.0016)\) & \(0.1516\;(0.0016)\)\\
4096 & \(0.4551\;(0.0013)\) & \(0.4541\;(0.0013)\) & \(0.1545\;(0.0017)\) & \(0.1543\;(0.0016)\)\\
\midrule
Poisson limit & \(0.4548\) & \(0.4548\) & \(0.1526\) & \(0.1526\)\\
\bottomrule
\end{tabular}
\end{table}

At \(m=4096\), the relative Monte Carlo standard errors of the centered estimates are \(0.29\%\) and \(1.07\%\), respectively. The largest absolute fine-to-coarse change in the raw estimates is \(0.00042\). Both centered estimates at \(m=4096\) lie within two Monte Carlo standard errors of their Poisson limits.

\section{Finite formulas and original replay controls}

Two implementations compare the finite depth formula with direct recursive integration over uniformly chosen cuts in rational arithmetic. All \(3{,}451\) cases agree. These cases cover each subset of sizes two through six of \(\{0,1,3,4,7,10,12\}\) and half-unit queries from \(-1\) through \(13\). The implementations also check the harmonic endpoint and the limiting mean bracket correction.

The original population simulations use the uniform density and \(2(1+t)/3\) on \([0,1]\), bag sizes \(128,512,2048\), and \(8{,}192\) replicates per size and density. At \(m=2048\) the maximum absolute error from the corrected mean across the query grid is \(0.02650\) cuts when sorting observations and \(0.00821\) cuts with the independent exponential-spacing implementation. The estimated mean insertion shifts at the interior queries lie between \(0.49386\) and \(0.50207\). The mean absolute uniform bracket contrast at \(x=1/4\) is \(0.5690\) and \(0.5817\) in the two implementations. At \(m=512\), forcing an observation at \(x=1/2\) or at \(x+0.02/m\) raises the mean at either location. The shifts are about \(0.493\) cuts in the first implementation and \(0.495\) in the second.

\begin{table}[htbp]
\centering
\small
\caption{Original covariance checks for \(m=128\) and uniform \(x=1/2\). The \(N=128\) row estimates \(\Var(D_m)\), since both bags are the full pool. Parentheses give Monte Carlo standard errors.}
\label{tab:original}
\begin{tabular}{@{}rrr@{}}
\toprule
\(N\) & Direct pool sampling & Hypergeometric overlap\\
\midrule
128  & \(0.5049\;(0.0056)\) & \(0.5194\;(0.0058)\)\\
256  & \(0.2211\;(0.0044)\) & \(0.2243\;(0.0045)\)\\
1024 & \(0.0525\;(0.0040)\) & \(0.0520\;(0.0041)\)\\
\bottomrule
\end{tabular}
\end{table}

Each shared-pool covariance in Table~\ref{tab:original} uses \(16{,}384\) pairs of bag scores. Independent-pool control covariances are within \(0.0031\) of zero in all cases. The code and report checksums accompany this supplement.

\end{document}